\documentclass[runningheads]{llncs}
\usepackage[T1]{fontenc}
\usepackage{graphicx}
\usepackage{amsmath}

\usepackage{cite}
\usepackage{booktabs}
\usepackage{bm}
\usepackage{multirow}
\begin{document}
\title{Are Large Language Models More Honest in Their Probabilistic or Verbalized Confidence?}
%
%
\author{Shiyu Ni\inst{1,2} \and
Keping Bi\inst{1,2} \and
Lulu Yu\inst{1,2} \and
Jiafeng Guo\inst{1,2}\thanks{Jiafeng Guo is the corresponding author}}
\authorrunning{Shiyu Ni et al.}
%
\institute{CAS Key Lab of Network Data Science and Technology, ICT, CAS \and
University of Chinese Academy of Sciences \email{\{nishiyu23z,bikeping,yululu23s,guojiafeng\}@ict.ac.cn}}

\maketitle              
\begin{abstract}
Large language models (LLMs) have been found to produce hallucinations when the question exceeds their internal knowledge boundaries. A reliable model should have a clear perception of its knowledge boundaries, providing correct answers within its scope and refusing to answer when it lacks knowledge. Existing research on LLMs' perception of their knowledge boundaries typically uses either the probability of the generated tokens or the verbalized confidence as the model's confidence in its response. However, these studies overlook the differences and connections between the two. In this paper, we conduct a comprehensive analysis and comparison of LLMs' probabilistic perception and verbalized perception of their factual knowledge boundaries. First, we investigate the pros and cons of these two perceptions. Then, we study how they change under questions of varying frequencies. Finally, we measure the correlation between LLMs' probabilistic confidence and verbalized confidence. Experimental results show that 1) LLMs' probabilistic perception is generally more accurate than verbalized perception but requires an in-domain validation set to adjust the confidence threshold. 2) Both perceptions perform better on less frequent questions. 3) It is challenging for LLMs to accurately express their internal confidence in natural language.

\end{abstract}
\section{Introduction}
Recently, large language models (LLMs) have demonstrated remarkable performance across various NLP tasks~\cite{ouyang2022training, brown2020language, fan2024reformatted}.  Despite their impressive capabilities, LLMs have several significant limitations. One critical issue is that LLMs can produce hallucinations, generating factually incorrect answers that appear accurate, primarily occurring when the question exceeds the model's internal knowledge boundaries~\cite{yin2024benchmarking}. 
A reliable system should provide correct answers when it knows the answer and refuses to answer when it does not, rather than fabricating unreliable responses, which is especially important in areas such as safety and healthcare. This requires the model to have a clear understanding of its knowledge boundaries, knowing what it knows and what it does not know. 

A model with a clear perception of its knowledge boundaries is not only more reliable but can also aid downstream tasks. For example, it can help retrieval augmentation (RA) where RA can be triggered only when the model expresses uncertainty about its answers to enhance efficiency and effectiveness, which we call adaptive retrieval augmentation. This is because retrieval augmentation incurs additional overhead and the quality of retrieved documents cannot be guaranteed, potentially misleading the model instead.

Existing research on the model's perception of its knowledge boundaries mainly involves two types of confidence: probabilistic confidence ~\cite{guo2017calibration, desai2020calibration, jiang2021can, kadavath2022language, si2022prompting} where they use the probability of the generated tokens as the model's confidence and verbalized confidence where LLMs are taught to express their confidence in words~\cite{lin2022teaching, yin2023large, tian2023just, xiong2023can, yang2023alignment, ni2024llms}. These represent the model's probabilistic and verbalized perceptions of its knowledge boundaries. However, these works only explore these perspectives separately, overlooking their differences and connections. \looseness=-1

In this paper, we investigate LLMs' probabilistic perception and verbalized perception of their factual knowledge boundaries, analyzing the differences and correlations between them. Specifically, we try to answer three research questions. \textbf{RQ1: What are the pros and cons of these two perceptions?} Inspired by the previous finding that LLMs can generate more accurate answers for more common questions~\cite{mallen2022not}, we also wonder \textbf{RQ2: How do these two perceptions change under questions of varying frequencies?} In addition to exploring the differences between these two perceptions, we also study their correlations, so the last research question is \textbf{RQ3: Can LLMs accurately express their internal confidence in natural language?}.

To answer \textbf{RQ1}, we choose four widely used LLMs and conduct experiments on the representative factual QA benchmark, i.e., Natural Questions (NQ)~\cite{kwiatkowski2019natural}. Experimental results indicate that LLMs' probabilistic perception of their knowledge boundaries is more accurate than their verbalized perception. However, probabilistic perception necessitates the use of an in-domain dataset to determine an appropriate confidence threshold for binarizing continuous probabilistic confidence. In contrast, LLMs' verbalized perception performs at a reasonable level without requiring additional setup.

To answer \textbf{RQ2}, we test two powerful black-box models on the Parent and Child dataset~\cite{berglund2023reversal} where questions in the Child dataset are less common than those in the Parent dataset. We find that both LLMs' probabilistic perception and verbalized perception of their knowledge boundaries perform better on the Child dataset than on the Parent dataset. This indicates that LLMs' perception levels decline on more familiar questions. Additionally, for less common questions, probabilistic perception outperforms verbalized perception by a greater margin.

To answer \textbf{RQ3}, we adopt two commonly used correlation coefficient methods: the Spearman correlation coefficient~\cite{hauke2011comparison} and the Kendall correlation coefficient~\cite{abdi2007kendall}. These methods are used to calculate the correlation between probabilistic confidence and verbalized confidence for four LLMs (the same models used in RQ1) on the NQ, Parent, and Child datasets. We show that, overall, LLMs' verbalized confidence is positively correlated with their probabilistic confidence. However, at a finer granularity, the correlation is weak and varies significantly across different datasets. Therefore, we conclude that it is challenging for LLMs to accurately express their internal confidence in natural language.

\section{Related Work}
Many studies have investigated deep neural models' perception of their knowledge boundaries, which can be primarily divided into two categories: probabilistic perception and verbalized perception.
\paragraph{\textbf{Probabilistic Perception}} This series of works~\cite{guo2017calibration, desai2020calibration, jiang2021can, kadavath2022language, si2022prompting} utilizes the probability of the generated tokens as the model's confidence in the answer, which we refer to as probabilistic confidence. It has been observed that deep neural networks tend to be overconfident, a problem that can be effectively mitigated by adjusting the generation temperature~\cite{guo2017calibration}. Subsequent research has examined the perception levels of pre-trained transformers. It has been found that BERT-style models are generally well-calibrated~\cite{desai2020calibration}, whereas generative language models lack this level of perception~\cite{jiang2021can}. More recent studies have explored LLMs' perception of their knowledge boundaries.  Kadavath et al.~\cite{kadavath2022language} and Si et al.~\cite{si2022prompting} demonstrate that using appropriate prompting techniques can make LLMs more reliable. 

\paragraph{\textbf{Verbalized Perception}}  With the development of LLMs, several studies have demonstrated that LLMs can express their confidence in words~\cite{lin2022teaching, yin2023large, tian2023just, xiong2023can, yang2023alignment, ni2024llms}, which we refer to as verbalized confidence. Lin et al.~\cite{lin2022teaching} first show that a model (GPT-3) can learn to express confidence about its answers in natural language. Then, Yin et al.~\cite{yin2023large} evaluate LLMs' self-knowledge by assessing whether they can identify unanswerable or unknowable questions. To enhance LLMs' ability to verbalize their confidence, some studies focus on prompting methods~\cite{tian2023just, xiong2023can} while Yang et al.~\cite{yang2023alignment} are dedicated to training methods.

These studies only explore LLMs' perception of their knowledge boundaries from either a probabilistic or verbalized perspective, overlooking the differences and connections between them, which we investigate in this paper.
\section{Preliminaries}
In this section, we will introduce our task and the basic experimental setup.
\subsection{Task Formulation}
\paragraph{\textbf{Open-Domain QA}.} The goal of open-domain QA is to ask the model $\mathcal{M}$ to provide an answer $a$ for a given question $q$. Unlike previous small-scale models~\cite{devlin2018bert} that rely on the retrieve-then-read pipeline~\cite{karpukhin2020dense, lewis2020retrieval, ni2023comparative}, where relevant external documents are first retrieved for the question $q$ and then the model extracts the correct answer from these documents, LLMs can answer the question directly based on their internal knowledge. We instruct LLMs to answer using prompt $p$ and the format can be described as:
\begin{equation}
    a = f_{\text{LLM}}(q, p)
\end{equation}
\paragraph{\textbf{Confidence Elicitation}.} Instead of only obtaining the answer, we also expect the model to express its confidence $c$ in the answer. We focus on two types of confidence: verbalized confidence and probabilistic confidence.

\textbf{Verbalized Confidence.} LLMs are found to have the power to express their confidence in words which we refer to as verbalized confidence $c_{verb}$. We use $\hat{p}$ to instruct the model to generate the answer along with its verbalized confidence and the format is:
\begin{equation}
    (a, c_{verb}) = f_{\text{LLM}}(q, \hat{p})
\end{equation}
where $c_{verb}=1$ indicates the model is confident in its answer while $c_{verb}=0$ means the opposite.

\textbf{Probabilistic Confidence.} Perplexity can reflect the model's internal degree of certainty in the answer which we refer to as probabilistic confidence $c_{prob}$. For an answer $a$ consisting of $n$ tokens $\{a_1, a_2, \cdots a_n\}$, $c_{prob}$ is computed as:
\begin{equation}
    c_{prob} = -\frac{1}{n}\sum_{i=1}^n\log{P(a_i|a_{<i})}
\end{equation}
where a lower $c_{prob}$ implies that the model is more confident in the answer.

\subsection{Experimental Setup}
\paragraph{\textbf{Prompts.}} To facilitate the model answering the question and expressing its confidence in the answer, we use prompt $\hat{p}$=\textit{``Answer the following question based on your internal knowledge with one or few words. If you are sure the answer is accurate and correct, please say certain after the answer. If you are not confident with the answer, please say uncertain. Question: \{question\}. Answer:"} where \textit{\{question\}} is the placeholder for the question $q$.
\paragraph{\textbf{Models.}} We conduct experiments on two representative open-source models (Llama2-7B-Chat and Mistral-7B-Instruct-v0.2) and two widely used black-box models that can return the probability of the generated tokens, including ChatGPT (gpt-3.5-turbo-1106) and GPT-Instruct (gpt-3.5-turbo-instruct). For all the models, we set temperature=1 to obtain the raw probabilities.

\section{Evaluating LLMs' Probabilistic and Verbalized Perceptions of Their Knowledge Boundaries\label{sec: performance on NQ}}
In this section, we investigate the performance of LLMs' probabilistic and verbalized perceptions of their knowledge boundaries and try to answer \textbf{RQ1}.  
\subsection{Exprimental Setup}
\paragraph{\textbf{Datasets.}} We conduct experiments on a widely used open-domain QA dataset, Natural Questions (NQ)~\cite{kwiatkowski2019natural}. NQ is constructed using Google Search queries with annotated short or long answers related to factual knowledge. For our experiments, we use only questions with short answers from the test set and treat these short answers as labels. We randomly sample 20\% of the data as the validation set, and report results on the remaining data.

\paragraph{\textbf{Metrics.}} Following previous research~\cite{ni2024llms}, we use \textbf{Alignment}, \textbf{Overconfidence}, and \textbf{Conservativeness} to measure LLMs' perception of their knowledge boundaries. \textbf{Accuracy} (acc for short) is employed to represent the QA performance, where a response is deemed correct if it contains the ground-truth label. \textbf{Uncertain rate} is computed as the proportion of samples where the model expresses uncertainty and is used to represent the model's uncertainty level.

In view of verbalized perception, $Alignment_{verb}$ is computed by the proportion of samples where LLMs' confidence matches the correctness of the response (i.e., $c_{verb}$=$acc$). $Overconfidence_{verb}$ is the proportion of samples where the model is confident but the response is incorrect (i.e., $c_{verb}=1, acc=0$), and $Conservativenes_{verb}$ is used to measure the proportion of samples where the model expresses uncertainty but the response is correct (i.e., $c_{verb}=0, acc=1$). 

Unlike verbalized confidence, probabilistic confidence is a continuous value and cannot be directly matched with binary accuracy. Therefore, we set a threshold $\lambda$ to binarize probabilistic confidence and the format is:
\begin{equation}
    c_{prob} = \begin{cases} 1, & \text { if } c_{prob} \leq \lambda \\ 
                             0, & \text { if } c_{prob} > \lambda \end{cases}
\end{equation}

Then, similar to the metrics for verbalized perception, we compute $Alignment_{prob}$, $Overconfidence_{prob}$, and $Conservativeness_{prob}$. We use the confidence threshold which achieves the optimal $Alignment_{prob}$ on the validation set as $\lambda$.

\subsection{Results and Analysis}

\begin{table}[h]
\centering
  \caption{LLMs' probabilistic and verbalized perceptions of their knowledge boundaries of LLMs on NQ. Bold denotes the highest score for each model. Unc., Conserv., and Overconf. stand for Uncertain rate, Conservativeness, and Overconfidence, respectively.}
  \scalebox{1.0}{\begin{tabular}{ccccccc}
    \toprule
    \textbf{Model} & \textbf{Acc} & \textbf{Strategy}  & \textbf{Unc.}& \textbf{Alignment} & \textbf{Overconf.} & \textbf{Conserv.} \\
    \midrule
    \multirow{2}{*}{\textbf{Llama2}} &  \multirow{2}{*}{0.2957} & Verb. & 0.1894 & 0.4512 & \textbf{0.5319} & 0.0170   \\
     &  &  Prob. & \textbf{0.8764} & \textbf{0.7254}  & 0.0512 & \textbf{0.2233} \\
    \midrule
     \multirow{2}{*}{\textbf{Mistral}} &  \multirow{2}{*}{0.2985} & Verb. & 0.4848 & 0.6260 &  \textbf{0.2954} & 0.0786  \\
     &  &  Prob. & \textbf{0.9034} & \textbf{0.7185}  & 0.0398 & \textbf{0.2417} \\
    \midrule
     \multirow{2}{*}{\textbf{GPT-Instruct}} &  \multirow{2}{*}{0.4021} & Verb. & 0.1868 & 0.5182 & \textbf{0.4464} & 0.0354  \\
     &  &  Prob. & \textbf{0.6891} & \textbf{0.6551} & 0.1269 & \textbf{0.2180} \\
    \midrule
    \multirow{2}{*}{\textbf{ChatGPT}} & \multirow{2}{*}{0.4229}  & Verb. & 0.2111 & 0.5252 & \textbf{0.4204} & 0.0554  \\
     &  &  Prob. & \textbf{0.6443} & \textbf{0.6741} & 0.1294 & \textbf{0.1985} \\
     
    \bottomrule
  \end{tabular}}

  \label{tab:nq_compare_results}
\end{table}

\begin{figure}[htbp]
  \centering
    \includegraphics[width=0.9\textwidth]{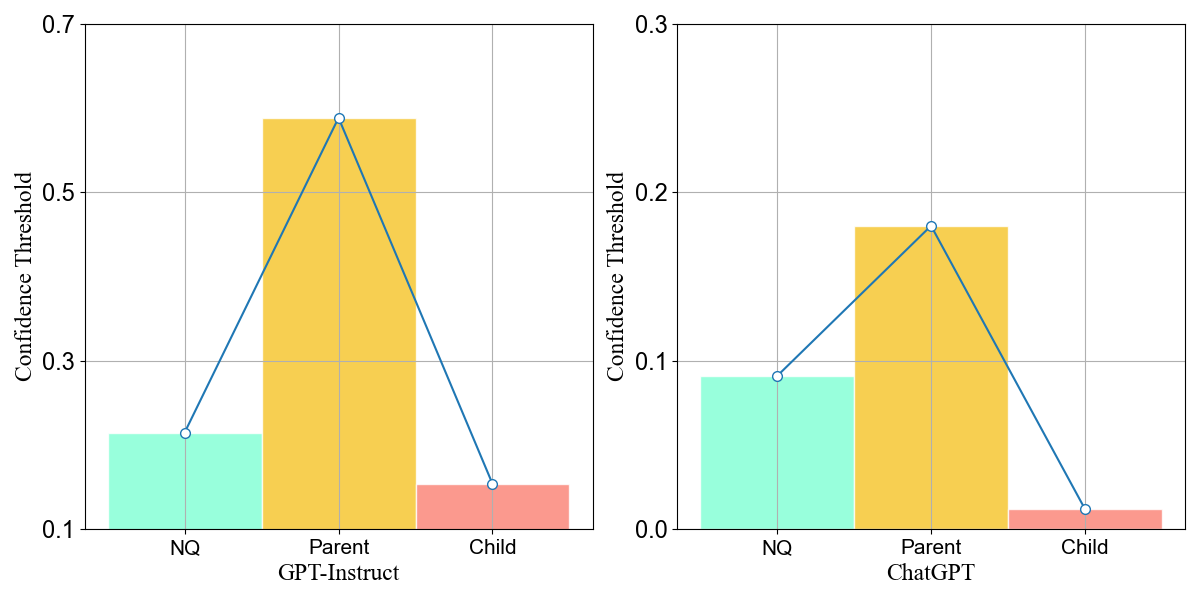}
  \caption{The best threshold $\lambda$ for GPT-Instruct and ChatGPT on each dataset.}
  \label{fig:ppl_conf_thre}
\end{figure}
The results of LLMs' QA performance and probabilistic and verbalized perceptions of their knowledge boundaries are shown in Table \ref{tab:nq_compare_results}. We observe that: 
1) When expressing confidence in words, LLMs are not well-calibrated and tend to be overconfident which is consistent with the previous findings~\cite{ni2024llms}. 
2) The probabilistic confidence is consistently much lower than the verbalized confidence, and the probabilistic alignment is significantly higher than the verbalized alignment across all models. This indicates that, compared to judging the correctness of an answer in words, LLMs have a better probabilistic perception of their knowledge boundaries. The possible reason may be that, when expressing confidence in words, LLMs do not have access to the probability distribution of the generated answer, which can be a useful signal representing the correctness of the answer.
3) A good probabilistic perception requires an additional in-domain dataset to select an appropriate threshold. Figure~\ref{fig:ppl_conf_thre} shows that the best probabilistic confidence of each model varies significantly across different datasets.

\section{Effects of Question Frequency \label{sec: effects of question frequency}}
LLMs often achieve better QA performance on common questions compared to unfamiliar ones~\cite{mallen2022not}. In this section, we investigate the effects of question frequency on LLMs' perception of their knowledge boundaries and answer \textbf{RQ2}.

\begin{table}[htbp]
\centering
  \caption{LLMs' probabilistic and verbalized perceptions of their knowledge boundaries on Parent and Child datasets. Bold denotes the highest score for each model. Unc., Conserv., and Overconf. stand for Uncertain rate, Conservativeness, and Overconfidence respectively.}
  \scalebox{1.0}{\begin{tabular}{cccccccc}
    \toprule
    \textbf{Model}  & \textbf{Dataset}  & \textbf{Acc} & \textbf{Strategy}  & \textbf{Unc.}& \textbf{Alignment} & \textbf{Overconf.} & \textbf{Conserv.} \\
    \midrule
     \multirow{4}{*}{\textbf{GPT-Instruct}} & \multirow{2}{*}{Parent} &  \multirow{2}{*}{\textbf{0.5475}} & Verb. & 0.3642 & 0.6424 & 0.2230 & \textbf{0.1346}  \\
     & & &  Prob. & 0.2114 & 0.6846  & \textbf{0.2783} & 0.0372 \\
     \cmidrule(lr){2-8}
     & \multirow{2}{*}{Child} & \multirow{2}{*}{0.1540} & Verb. & 0.6018 & 0.7243 & 0.2599 & 0.0157 \\
     & &  & Prob. & \textbf{0.8609} & \textbf{0.8593}  & 0.0629 & 0.0778\\
     \midrule
      \multirow{4}{*}{\textbf{ChatGPT}} & \multirow{2}{*}{Parent} &  \multirow{2}{*}{\textbf{0.5778}} & Verb. & 0.2288 & 0.6198 & 0.2868 & 0.0934  \\
     & & &  Prob. & 0.3877 & 0.7670 & 0.1337 & \textbf{0.0992} \\
     \cmidrule(lr){2-8}
     & \multirow{2}{*}{Child} & \multirow{2}{*}{0.1322} & Verb. & 0.5632 & 0.6803 & \textbf{0.3121} & 0.0075\\
     & &  & Prob. & \textbf{0.9339} & \textbf{0.8820} & 0.0259 & 0.0921\\
    \bottomrule
  \end{tabular}}

  \label{tab:parent_child_res}
\end{table}
\subsection{Experimental Setup}
\paragraph{\textbf{Datasets.}} The Parent-Child dataset is a collection of facts about actual celebrities and their parents, presented in the form ``A's parent is B" and ``B's child is A" where A is the name of the celebrity and B is the name of A's parent~\cite{berglund2023reversal}. LLMs are often more familiar with questions asking about the names of famous people's parents because these are more likely to appear in training corpora. We collect questions in the form "Who is A's mother/father" and name this dataset "Parent." Correspondingly, questions in the form "Name a child of B" are named "Child." Each dataset contains 1513 question-answer pairs. The dataset splitting strategy, metrics, and other parameters are the same as those in Section~\S\ref{sec: performance on NQ}.

\paragraph{\textbf{Models}} We find that Llama and Mistral often refuse to answer questions containing names. So, we conduct experiments only on GPT-Instruct and ChatGPT.

\subsection{Results and Analysis}
Table~\ref{tab:parent_child_res} shows the QA performance of GPT-Instruct and ChatGPT on questions of different frequencies, along with the models' perception of their knowledge boundaries. We find that: 
1) LLMs achieve better QA performance which aligns with the previous findings in~\cite{berglund2023reversal} and are more confident in the parent dataset compared to the child dataset.
2) Both verbalized and probabilistic alignment are higher on the child dataset. This indicates that LLMs have a better perception of their knowledge boundaries on less common questions rather than more familiar ones. 
3) From common questions to unfamiliar ones, probabilistic alignment demonstrates a greater increase compared to verbalized alignment. The reason is that probabilistic confidence drops to a very low level on the child dataset which mitigates the level of overconfidence. It shows that these models have a very clear probabilistic understanding of what they do not know on the unfamiliar questions. At the same time, they also maintain the perception level of what they know. In view of verbalized perception, LLMs have a more accurate judgment of what they know. However, they are still overconfident which is the primary reason for the unsatisfactory perception of knowledge boundaries.

\section{The Correlation Between LLMs' Probabilistic Confidence and Verbalized Confidence}
 In this section, we study the correlation between LLMs' probabilistic confidence and verbalized confidence to answer \textbf{RQ3}.

\subsection{Experimental Setup}
For a probabilistic confidence list $\mathcal{C}_p=\{c_{p}^1, c_p^2, \cdots c_p^n\}$ and a verbalized confidence list $\mathcal{C}_v=\{c_v^1, c_v^2, \cdots c_v^n\}$, we utilize two commonly used correlation coefficients: Spearman~\cite{hauke2011comparison} and Kendall~\cite{abdi2007kendall} correlation coefficients to measure their correlations.

\paragraph{\textbf{Spearman Coefficient.}}
Spearman's rank correlation coefficient uses the rank of each value in the lists $\mathcal{C}_p$ and $\mathcal{C}_v$ to measure their correlation and the formula is:
\begin{equation}
    \rho_s=1-\frac{6 \sum d_i^2}{n\left(n^2-1\right)}
\end{equation}
where $d_i = rank_{c_p^i} - rank_{c_v^i}$ and $n$ is data count. The value of $\rho_s$ ranges from [-1, 1], where a larger absolute value indicates a stronger correlation. 1 represents a perfect positive correlation, while -1 represents a perfect negative correlation.
\paragraph{\textbf{Kendall Coefficient.}}
Kendall's rank correlation coefficient is defined based on the concepts of concordant pairs and discordant pairs. A concordant pair or discordant pair refers to a pair where the relative ordering of the two variables is consistent (e.g., $c_p^i>c_p^j$ and $c_v^i>c_v^j$) or not (e.g., $c_p^i>c_p^j$ and $c_v^i<c_v^j$). The format is:
\begin{equation}
    \rho_k=\frac{e-f}{e+f}=\frac{e-f}{\frac{1}{2} \cdot n \cdot(n-1)}
\end{equation}
where $e$ is the count of concordant pairs, $f$ is the count of discordant pairs, and $n$ is the data count. Similar to $\rho_s$, the value of $\rho_k$ also ranges from [-1, 1], where 1 represents a perfect positive correlation, while -1 represents the verse vice.


\paragraph{\textbf{Mode.}} We calculate the correlation coefficients in two modes. 1) \textbf{Vanilla}: We take LLMs' probabilistic confidence and verbalized confidence for all the data as $\mathcal{C}_p$ and $\mathcal{C}_v$. 2) \textbf{Bin-k}. To mitigate the influence of the order of individual samples and estimate the overall trend, we sort all the data in ascending order based on the probabilistic confidence and divide them into $k$ bins with the same length. The probabilistic confidence and verbalized confidence of each bin are the average values of the data within that bin. This yields lists $\mathcal{C}_p$ and $\mathcal{C}_v$, each of length $k$. In this paper, We set $k$ to 10.

\begin{table}[htbp]
\centering
  \caption{Correlation coefficients between LLMs' verbalized confidence and probabilistic confidence. Bold denotes the highest score on each dataset.}
  \scalebox{1.0}{\begin{tabular}{cccccccccc}
    \toprule
     & & \textbf{Llama} & \textbf{Mistral} & \multicolumn{3}{c}{\textbf{GPT-Instruct}} & \multicolumn{3}{c}{\textbf{ChatGPT}} \\
     \cmidrule(lr){5-7} \cmidrule(lr){8-10}
      & & \textbf{NQ} & \textbf{NQ}  & \textbf{NQ} & \textbf{Parent} & \textbf{Child} & \textbf{NQ} & \textbf{Parent} & \textbf{Child} \\
    \midrule

    \multirow{2}{*}{\textbf{Vanilla}} & Spearman & 0.24	& \textbf{0.37} & 0.23 & 0.13 & \textbf{0.38} & 0.22 & \textbf{0.2} & 0.28 \\
    & Kendall & 0.2	& \textbf{0.3} & 0.19 & 0.1 & \textbf{0.31} & 0.18 & \textbf{0.16}	& 0.23 \\
    \midrule

    \multirow{2}{*}{\textbf{Bin-10}} & Spearman  & 0.75	& 0.81 & 0.9 & 0.12	& \textbf{0.92} & \textbf{0.95} & \textbf{0.88} & 0.48 \\
    & Kendall & 0.73 & 0.73	& 0.78 & 0 & \textbf{0.85} & \textbf{0.87} & \textbf{0.78}	& 0.33 \\

    \bottomrule
  \end{tabular}}

  \label{tab:ppl_conf_correlation}
\end{table}
\subsection{Results and Analysis}

\begin{figure}[htbp]
  \centering
    \includegraphics[width=\textwidth]{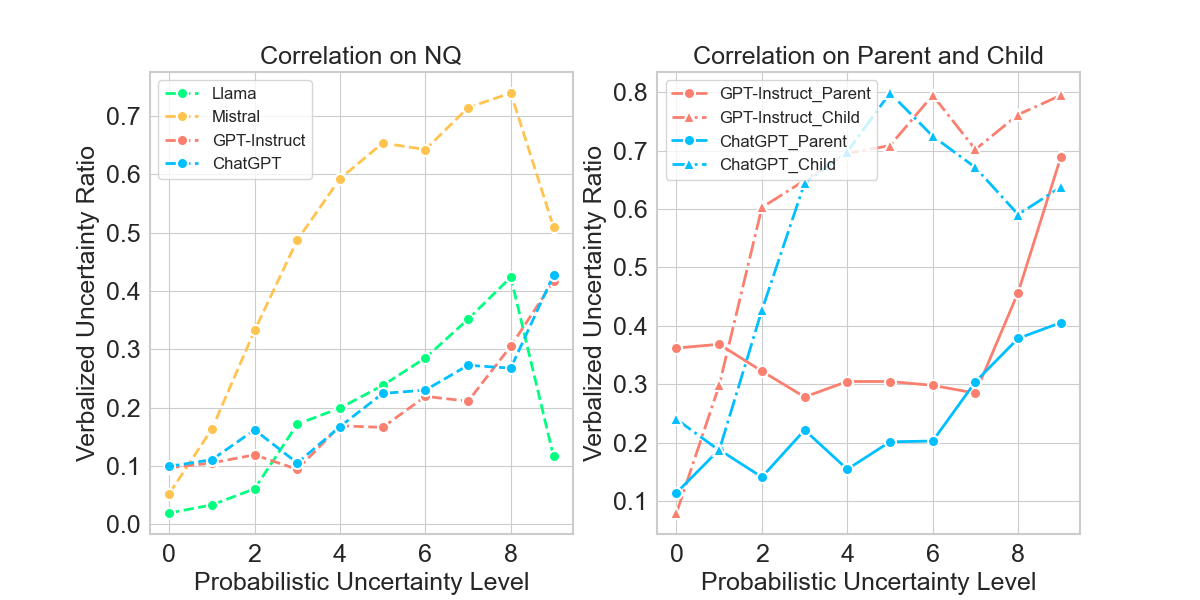}
  \caption{Correlation between LLMs' probabilistic confidence and verbalized confidence. A higher uncertainty level means the model is less confident in its answer.}
  \label{fig:ppl_conf_correlation}
\end{figure}

The results can be seen in Table~\ref{tab:ppl_conf_correlation} and we visualize the changes in verbalized confidence with probabilistic confidence under the Bin-10 mode, as shown in Figure \ref{fig:ppl_conf_correlation}. We conclude that: 
1) In Vanilla mode, the correlation coefficient is small, but it is much higher in Bin-10 mode (except for ChatGPT on the Parent dataset). This indicates that the correlation between the model's probabilistic confidence and verbalized confidence is relatively low although there is an overall trend showing that verbalized uncertainty increases as probabilistic uncertainty increases. 
2) The correlation varies significantly across different datasets for the same model. For instance, probabilistic confidence and verbalized confidence of GPT-instruct show a clear overall trend on the child dataset, whereas, on the parent dataset, they are almost entirely unrelated. Therefore, it is challenging for LLMs to accurately express their internal confidence in words. 

\subsubsection{Correlation Between LLMs' Confidence and Their QA Performance.} To more intuitively describe LLMs' perception of their knowledge boundaries, we investigate the correlation between probabilistic confidence and QA performance, as well as between verbalized confidence and QA performance, without binarizing the probabilistic confidence using a threshold. We calculate these correlations using Spearman and Kendall coefficients in both Vanilla and Bin-10 modes. The results are shown in Table~\ref{tab:coefficient}. We observe that: 
1) Both LLMs' probabilistic confidence and verbalized confidence are positively correlated with their QA performance, with probabilistic confidence generally showing a stronger correlation with QA performance than verbalized confidence. This indicates that LLMs have a better probabilistic perception of their knowledge boundaries even without a precise threshold, which aligns with the conclusion in Section~\S\ref{sec: performance on NQ}. 
2) Both correlations are very strong in the Bin-10 but relatively weak in the vanilla mode, representing that LLMs have a less nuanced but overall clear perception of their knowledge boundaries.

\begin{table}[htbp]
\centering
  \caption{Correlation coefficients between LLMs' confidence and their QA performance. Bold denotes the highest score for each correlation coefficient. Prob. and Verb. stand for probabilistic confidence and verbalized confidence, respectively.}
  \scalebox{1.0}{\begin{tabular}{ccccccccccc}
    \toprule
     & & & \textbf{Llama} & \textbf{Mistral} & \multicolumn{3}{c}{\textbf{GPT-Instruct}} & \multicolumn{3}{c}{\textbf{ChatGPT}} \\
     \cmidrule(lr){6-8} \cmidrule(lr){9-11}
     & & & \textbf{NQ} & \textbf{NQ}  & \textbf{NQ} & \textbf{Parent} & \textbf{Child} & \textbf{NQ} & \textbf{Parent} & \textbf{Child} \\
    \midrule

    \multirow{4}{*}{\textbf{Vanilla}} & \multirow{2}{*}{Spearman} & Prob. & \textbf{0.31} & 0.23 & \textbf{0.34} & \textbf{0.44} & 0.41 & \textbf{0.35} & \textbf{0.58} & \textbf{0.42} \\
    & & Verb. & 0.22 & \textbf{0.28} & 0.22	& 0.26	& \textbf{0.44}	& 0.19	& 0.18	& 0.37 \\
    
    \cmidrule(lr){2-11}

    & \multirow{2}{*}{Kendall} & Prob. & \textbf{0.25} & 0.19 & \textbf{0.28} & \textbf{0.36} & 0.33 & \textbf{0.29} & \textbf{0.47} & 0.35 \\
    & & Verb. & 0.22 & \textbf{0.28} & 0.22 & 0.26 & \textbf{0.44}	& 0.19	& 0.18	& \textbf{0.37} \\
    \midrule

    \multirow{4}{*}{\textbf{Bin-10}} & \multirow{2}{*}{Spearman} & Prob. & \textbf{0.99} & \textbf{0.99} & \textbf{1} & \textbf{0.99} & \textbf{0.98} & \textbf{1} & \textbf{0.98} & 0.89 \\
    & & Verb. & 0.7 & 0.91 & 0.7	& 0.88	& 0.85	& 0.45	& 0.81	& \textbf{0.92} \\
    
    \cmidrule(lr){2-11}

    & \multirow{2}{*}{Kendall} & Prob. & \textbf{0.96} & \textbf{0.96} & \textbf{1} & \textbf{0.96}	& \textbf{0.94}	& \textbf{1} & \textbf{0.94} & 0.81 \\
    & & Verb. & 0.62 & 0.81 & 0.62 & 0.78 & 0.73 & 0.35 & 0.71 & \textbf{0.85} \\

    \bottomrule
  \end{tabular}}

  \label{tab:coefficient}
\end{table}

\section{Conclusion}
In this paper, we conduct a comprehensive analysis and comparison of LLMs’ probabilistic and verbalized perceptions of their factual knowledge boundaries. Specifically, we focus on answering three research questions: RQ1: What are the pros and cons of these two perceptions; RQ2: How do these two perceptions change under questions of varying frequencies; and RQ3: Can LLMs accurately
express their internal confidence in natural language. We conduct extensive experiments on four commonly used LLMs and three open-source datasets and find that 1) LLMs’ probabilistic perception is generally more accurate than verbalized perception but requires an in-domain validation set to adjust the confidence threshold. 2) Both perceptions perform better on less frequent questions and probabilistic perception outperforms verbalized perception by a greater margin on these questions. 3) It is challenging for LLMs to accurately express their internal confidence in natural language.

\bibliographystyle{splncs04}
\bibliography{ref}

\end{document}